\documentclass{article}

\usepackage{arxiv}

\usepackage[utf8]{inputenc} 
\usepackage[T1]{fontenc}    
\usepackage{hyperref}       
\usepackage{url}            
\usepackage{booktabs}       
\usepackage{amsfonts}       
\usepackage{nicefrac}       
\usepackage{microtype}      
\usepackage{lipsum}
\usepackage{algorithmic}
\usepackage{algorithm}
\usepackage{graphicx}
\graphicspath{ {./img/} }

\title{When deep learning models on GPU can be accelerated by taking advantage of unstructured sparsity}

\author{
Marcin Pietro\'n \\
  Department of of Computer Science\\
  AGH University of Science and Technology\\
  Cracow, Poland \\
  \texttt{pietron@agh.edu.pl} \\
   \And
  Dominik \.{Z}urek \\
  Department of Computer Science\\
  AGH University of Science and Technology\\
  Cracow, Poland \\
  \texttt{dzurek@agh.edu.pl} \\

}

\begin{document}
\maketitle

\begin{abstract}
This paper is focused on the improvement of the efficiency of sparse convolutional neural networks (CNNs) layers on graphic processing units (GPU). The Nvidia deep neural network (cuDnn) library provides one of the most effective implementations of deep learning (DL) algorithms for GPUs. GPUs are the most efficient and commonly used accelerators for deep learning computations. Modern CNN models need megabytes of coefficients and millions of MAC operations to perform convolution. One of the most common techniques for compressing CNN models is weight pruning and quantization. There are two main types of pruning: structural (based on removing whole weight kernels) and non-structural (removing individual weights). The first enables much easier acceleration on many type of accelerators, but with this type it is difficult to achieve a sparsity level and accuracy as high as that obtained with the second type. Non-structural pruning with retraining can generate a matrix-weight up to $\sim90\%$ or more of sparsity in some deep CNN models.  This work shows when it is worth using a direct sparse operation to speed-up the calculation of the convolution layers. In the next stage the linear and non-linear quantization is performed for further cycles and memory reduction. This work presents the impact of using reduced precision on time efficiency.
\end{abstract}


\section{Introduction}
Deep convolutional neural networks (CNNs) achieve outstanding results in various artificial intelligence tasks including image classification \cite{cnn_image_classification_1}\cite{cnn_image_classification_2}, object detection \cite{cnn_object_detection_1}, semantic segmentation and natural language processing  \cite{kim-2014-convolutional}\cite{cnn_text_quant_1}\cite{cnn_nlp_1}. Recent CNN neutral networks consist of dozens of convolution layers and a few fully connected layers. Neural networks for conducting the training process on large benchmark datasets need different accelerators such as multi-core processors, GPGPUs or other dedicated hardware accelerators. Over the years, scientists have been looking for methods to accelerate the calculations of the convolution operation. The direct convolution algorithm to perform convolutions requires $N^2$ multiplications and \textit{N(N-1)} additions where \textit{N} is the size of the input. For the same input the Fast Fourier Transform (FFT) method reduces operation complexity to \textit{O(N$log_{2}$(N))} \cite{fft_improvment}. The Winograd algorithm is suitable for small fixed-size kernels and requires 2.25 times fewer multiplications than direct convolution \cite{winograd_lavin}. The convolution operation can be realised by matrix-multiplication  \cite{conv_mm}, especially on the GPGPU which is highly tuned for performing this operation \cite{gmmem_gpu}. The GPGPU remains one of the most efficient and commonly used hardware accelerators. The NVIDIA deep neural network library (cuDNN)\footnote{https://developer.nvidia.com/cudnn} performs convolution with different algorithms (Winograd, FFT, GEMM) depending upon filter size, batch size and data representation. Apart from choosing different algorithms for speeding up convolution there are some other methods based on complexity and memory footprint reduction. Some CNN models for image processing or natural language processing can be heavily pruned. The effect of this process is that they very often contain zero values more than 80\% of weights. Depending upon the level of sparsity, it can be worth performing the convolution through the application of the direct sparse convolution method proposed by \textit{Chen} \cite{chen2018escoin}. The paper concentrates on two aspects. The first is about methods for complexity reduction. It explains pruning and quantization methodologies and theirs results. The second is focused on investigating when it is worth using sparse operations, instead of using dedicated NVIDIA libraries to perform the convolution layer on the GPGPU. As the main optimization strategy we propose the introduction of a unified sparse level for each of the output channels in each convolutional layer. The other optimization strategy is determining the most optimal number of thread blocks for each convolutional layer separately. The presented approach is optimized towards the optimal arrangement of the data in order to obtain acceleration with the direct convolution approach using the sparse format. These strategies are crucial for achieving peak performance. This work shows real examples of models where it is possible to obtain the appropriate level of sparsity so that acceleration using the presented algorithm vs cuDnn could be possible. These examples are well known CNN models used on smaller and less complex data sets. The high sparsity levels were obtained by the presented pruning approach. Apart from achieving a high sparsity and compression ratio, the accuracy levels were also improved. Finally, the impact on time efficiency of using half precision (FP-16) in a direct sparse convolution is explored. It is compared with cuDnn, where for 16-bit data representation, NVIDIA Tensor Cores specialised arithmetic units are used. In the presented work quantization is used in two variants: after pruning in the training process and on a pretrained model. Both linear and nonlinear approaches of quantization were applied. To our knowledge, this is the first work that shows the acceleration of the unstructured sparsity of weights compared to the dense approach using real models. 

\section{Related work}
Convolution complexity and efficiency optimization have recently become quite popular research subject. Jord\'a \textit{et al.} \cite{cudnn_comaprision} present the way in which the cuDnn library calculates convolution layers dependent upon parameter configurations and data representation. Lavin \textit{et al.} \cite{winograd_lavin} introduces Winograd convolution implementation which is based on minimal filtering algorithm. This approach for a small filter and batch size was 2.26 times faster than the previous version of cuDnn. Ad\'amek \textit{et al.} \cite{fft_improvment} proposes an FFT based convolution on GPGPU by the shared memory implementation of the overlap-and-save method, and for certain sizes, a 30\% speed increase was achieved in comparison to cuDnn. The direct sparse convolutions method was proposed in \cite{Liu_dircect_sparece_idea}. The authors used the CSR format to store the weights and perform the convolution operation by use of the sparse matrix multiplication. This approach achieved 3.1-7.3 times speed increase comparison to dense convolution in the AlexNet model, on Intel Atom, Xeon and Xeon Phi processors. \textit{Lu et al.} \cite{sparse_winograd_fpga} proposed FPGA's sparse convolution implementation which in VGG16 is almost three times faster than FPGA's dense implementation. The same type of convolution was applied on the GPU in \cite{chen2018escoin}, where the speed increase for AlexNet \cite{AlexNet_architecture}, GoogleLeNet \cite{GoogleLeNet_architecture} and ResNet \cite{ResNet_Architecture} models were respectively 1.74, 1.34 and 1.43 times that of the GEMM implementation in the CUBLAS\footnote{https://developer.nvidia.com/cublas} library. \textit{Zhu et al.} \cite{sparse_rnn} used sparse matrix operation in order to perform recurrent neural networks (RNN), where the data format of sparse persistent RNN is represented by the $\langle index, value \rangle$ pairs. The authors have proposed several optimization strategies for GPU implementation such as wide memory loads and bank-aware weight layouts. This approach for a hidden layer of size 1782 and density of 10\%  allows the following speed increases to be achieved: 7.3 times that of dense GEMM (cuDnn), 3.4 times that of sparse GEMM (cuSparse\footnote{https://docs.nvidia.com/cuda/cusparse}) and 1.8 times compared to dense persistent implementation (cuDnn). An important role in sparse convolution is played by weight pruning, which can produce a number of zero weights \cite{}. Information about the level of weight sparsity can be used after the pruning step in order to decide if it is worth running direct sparse implementation or cuDnn. The paper shows models and data sets on which it is possible to achieve a level of sparsity which can provide a better level of efficiency than cuDnn by using direct sparse convolution. In other research \cite{pietron2020retrain}, the authors prove that retraining with pruning can reduce the drop in accuracy caused by removing unimportant weights.
Pruning is one of the most popular solutions when it comes to memory compression and the acceleration of deep learning models \cite{pietron2020retrain}, \cite{frankle2019}, \cite{movement_pruning2020}. When it comes to accelerating models with pruning, dedicated accelerators are very often built (eg. based on FPGA) that can use unstructured pruning. Recently, there has been a lot of research on pruning. 
Some of the most popular approaches of pruning methods which incorporate retraining are: pruning without retraining with local search heuristics \cite{pietron2020retrain},\cite{motaz2020}, lottery ticket \cite{frankle2019}, movement pruning  \cite{movement_pruning2020} and \cite{10.1145/3007787.3001163}. In most of the mentioned works there is no real use of the results obtained from unstructured sparsity in the GPU. 
Many modern hardware accelerators support reduce bit precision arithmetic. Quantization is the next step by which it is possible to reduce workload and memory further. Many quantization approaches were applied for deep learning    
\cite{han2015learning} \cite{linda2016} \cite{gysel2016ristretto} using linear or nonlinear quantization, regularization modifications, clustering \cite{pietronCANDAR} and other techniques \cite{motaz2020}. 

\section{Convolutional neutral networks}
The typical convolutional layer in a feed-forward procedure calculates the convolution of the inputs which is represented by a batch of \textit{N} samples (images, time series etc.) with \textit{C} channels and size \textit{H$\times$W}, with a set of \textit{K} filters with \textit{C} channels and size \textit{R$\times$S}. The output product of convolution contains \textit{K} matrices with size \textit{E$\times$F}, where $ E = \frac{H + 2*padding - R}{stride} + 1$ and $F = \frac{W + 2*padding - S}{stride} + 1$. The set of parameters of a single convolution layer is a 4D array called a \textit{tensor}. When the kernel is marked as \textit{W} and the input is marked as \textit{I} the convolution of a single layer is given by the formula:
\begin{equation}
   Out_{n,i,j,k} = \sum_{c=0}^{C-1}\sum_{r=0}^{R-1}\sum_{s=0}^{S-1}W_{k,c,r,s}I_{n,c,i+r,j+s} 
\end{equation}
The result of the above formula is added to the bias parameter \textit{b} and the activation function is then applied \cite{activation_function_choosen}. The convolution layers are the most time-consuming operation in the CNN flow. For this reason, only these layers have been subjected to an acceleration in this paper. In our experiments the VGG-16 \cite{vgg-16_architecture}, CNN-non-static \cite{kim-2014-convolutional} some  1x1 layers from ResNet \cite{ResNet_Architecture} and DenseNet \cite{DenseNet} models were used as a benchmarks.

\section{Convolution algorithms on the GPGPU}
In order to perform forward convolution the cuDnn library always chooses the most effective algorithm, depending upon input, filter, batch size and data format. The graphics processing units (GPUs) are very effective particularly for accelerating large matrix products such as matrix-matrix multiplication and element-wise multiplication. For this reason, the most productive algorithms for performing convolution on GPGPUs firstly transforms the data to a form which allows performing the convolution through the application of these operations. The first most commonly used algorithm is general matrix multiply (GEMM). This method transforms the input and the filters into two matrices. Convolution is performed by the scalar product of the single row and the single column which is repeated for each input's column and all rows from transposed filters. This method is used when data are represented at half precision, in 1D, a 1$\times$1 convolution and when the number of channels is relatively small, which usually takes place for the first layers of most CNN's architecture. The second method is based on the Fast Fourier Transform (FFT). This method for transforming the convolution of two signals in one domain (e.g. time) is equivalent to the point-wise multiplication of their Fourier transform in the other domain (e.g. the frequent domain). After calculation the inverse Fast Fourier Transform is requested in order to return back to the time domain. In theory, FFT convolution is the most effective way to perform convolution for large filters like 5x5. Based on VGG-16 architecture, the cuDnn uses this method to perform convolution in the case of the input size being smaller than or equal to 58$\times$58 (from the sixth layer) and for a batch size higher than 32. The last method to perform convolution on GPGPU is Winograd. This method is based on the Chinese reminder theory (CRT)\cite{winograd1980arithmetic}. Thanks to introducing some transformations there is a reduction in the number of multiplications and an immediate increase in the number of required additions, which results in a faster computation.

\section{Pruning}

Very often, many deep learning models have a lot of redundant weights. Research in the exploration of pruning techniques has been recently showed that many cases of deep learning architecture can be compressed with high ratios. Several methodologies have already been tried. The most popular techniques are for example pruning with or without retraining \cite{pietron2020retrain}, incremental pruning or pruning with a constant sparsity \cite{frankle2019}, pruning with constant mask or dynamic, gradient-based pruning \cite{movement_pruning2020}.
The work focuses on the application of popular deep learning models to CIFAR100 and CIFAR10. In less complex data sets like CIFAR100 with a reduced number of classes (smaller than Imagenet) there is a higher probability of obtaining such sparsity levels that can give a faster solution than cuDnn. In addition, the work shows that not only such models can be accelerated on data sets of reduced complexity (e.g. with a smaller number of classes) but using the pruning approach, their accuracy can be significantly increase.

\subsection{Pruning approach}

The proposed pruning method is based on retraining. Pruning with retraining guarantees much better final sparsities. Algorithm \ref{alg:pruning} incorporates evolutionary techniques and rewinding during its execution. It makes it  possible to return to the values of nonzero weights which were before given iteration. The input parameters are: 
\begin{itemize}
    \item $acc\_threshold$ - threshold for accuracy acceptable accuracy changes 
    \item $iter\_nr$ - number of iterations of the algorithm 
    \item $batch\_nr$ - number of batches after which pruning configuration can be changed
    \item $pool\_size$ - number of solutions in a population
\end{itemize}

The algorithm starts from scratch with random initial weights and generates a pool of solutions (subnetworks) with random initial sparsities (line 1). In each iteration some subsets of layers are chosen for further pruning. This step helps to gather statistics about layer sensitivity and diagnoses which layer may be blocking learning. The batch training is then performed (it can be the whole epoch, it depends upon the algorithm settings).    

\begin{algorithm}
\begin{algorithmic}[1]
\REQUIRE{$acc\_threshold, iter\_nr, batch\_nr, pool\_size$}
\STATE{$generate\_pool\_of\_solutions$}
\FOR{$i < iter\_nr$}
\STATE{$choose\_subnetwork\_from\_pool$}
\STATE{$choose\_layers\_in\_model$}
\STATE{$train\_batches$}
\STATE{$compute\_grad\_statistics$}
\STATE{$compute\_accuracy$}
\IF{$improvement > acc\_threshold$}
\STATE{$write\_to\_pool\_if\_good\_enough$}
\STATE{$increment\_mask$}
\ELSE
\STATE{$mutation\_or\_crossover$}
\STATE{$rewinding$}
\ENDIF
\STATE{$alpha = check\_weights\_migration()$}
\STATE{$update\_sensitivity$}
\IF{$stagnation$}
\STATE{$differentiate\_solutions\_in\_pool$}
\ENDIF
\ENDFOR
\end{algorithmic}
\caption{Pruning - main scheme approach}
\label{alg:pruning}
\end{algorithm}

After the batch training gradient analysis is performed and the accuracy is measured on a validation set (line 6 and 7). The current solution is compared with others from the pool. If it is good enough it is written to the population set and its mask (sparsity) is incremented (line 9).
The mask in the presented algorithm is a dynamic structure which indicates which weights should be pruned. The mask is recomputed after each batch training (see eqs. \ref{alg:pruning2} and \ref{alg:pruning3}). The $wg_{n,i,j,k}$ coefficient for each weight is computed based on its absolute value and its current gradient value. $Alpha$ parameters define how important these factors are.

\begin{equation}
wg_{n,i,j,k} = alpha*grad_{n,i,j,k}+(1-alpha)*abs(w_{n,i,j,k})
\label{alg:pruning2}
\end{equation}


\begin{equation}
   mask_{n,i,j,k} = 1\; if \;wg_{n,i,j,k} \in max(wg, sparsity) \;else\; 0
\label{alg:pruning3}
\end{equation}

The mask increase is set on the basis of the sensitivity of pruned layers. If the progress in training the model is not satisfactory (line 8), rewinding and mutation or crossover is performed. Mutation is just a random sparsity change in the given layers. Crossover takes two random parents from the population and exchange theirs sparsity numbers in randomly chosen layers.

\begin{equation}
   w_{n,i,j,k} = w_{n,i,j,k}*mask_{n,i,j,k}  
   \label{alg:pruning1}
\end{equation}



In the next steps the $alpha$ parameter is computed on the basis of a weights migration statistics (from and into the mask). The sensitivity of the layers is updated, which indicates how the process of pruning specific layers affects the accuracy level. The last step helps to avoid stagnation in the algorithm. The population is divided into a specific number of clusters and only some constant representatives of each cluster stay in the population.    

\section{Quantization}

After the process of network distillation by the pruning process quantization can be performed as the next step of reducing model complexity. Quantization is the procedure of constraining values from a continuous set or more dense domain to a relatively discrete set. It is possible to define a general mapping from a set of floating-point data $x\in {\mathcal S}$ to fixed-point $q$ as follows (assuming signed representation):
\begin{equation}
	\label{eq:quant}
	q_{\textit{fxp}} = {\mathcal Q}(x_{\textit{flp}}) = \mu + \sigma \cdot \textit{round}(\sigma^{-1}\cdot (x-\mu)). 
\end{equation}

In our case $\mu=0$ and $\sigma = 2^{-{\bf frac\_bits}}$ where 
\begin{equation}
	{\bf int\_bits}=\textit{ceil}(\log_{2}(\max_{x\in{\mathcal S}} |x|)) 
\end{equation}
and ${\bf frac\_bits}={\bf total\_bits}-{\bf int\_bits}-1$.
The scaling factor $\sigma$ is essentially just a shift up or down. A drawback is that a great deal of precision may be lost if the distribution of the data set ${\mathcal S}$ is skewed by a large mean.
Yet another approach can define the number of integer and fractional bits to represent regions of a distribution that will represent a large percentage of the range.  
In these cases, there will be saturation of a small percentage of the data, such as outliers, through the quantization procedure which may or may not significantly affect the accuracy. To determine the effects of saturation one can experiment with different saturation levels. Therefore histogram analysis is used to analyze outliers and set the best levels of saturation for activation of quantization. 

The eq. \ref{eq:quant} can be adapted to mapping floating numbers to integers values:

\begin{equation}
	\label{eq:quant2}
	q_{\textit{int}} = {\mathcal Q}(x_{\textit{flp}}) = ceil((x -\mu) / ((max(X)-min(X)) \cdot \sigma^{-1}))
\end{equation}

The $\mu$ parameter can be set to $min(X)$ ($X$ is an input set of values to be quantized) or can have a value of zero. In the first case it is known as an asymmetric integer quantization (e.g. used in TensorFlow framework), in the second, it is called symmetric. In this work fixed point was applied.
To compare linear quantization results, a nonlinear technique based on clustering was implemented. This approach assigns weight values for given layer to a given number of centroids eq. \ref{eq:quant3}. After that each weight is assigned to a cluster centroid which it belongs to eq. \ref{eq:quant4}. The codebook of values is created. During inference/forward pass each original value is mapped to its reduced centroid representation (16 or 8 bit). This approach gives additional memory compression.

\begin{equation}
\label{eq:quant3}
C, W_c = clustering(W)
\end{equation}

\begin{equation}
\label{eq:quant4}
W_c = \{\forall w_c \in W_c, \exists c_i \in C: w_c = c_i \}
\end{equation}

\section{Building CSR weight format}
\label{weigh_pruning}
After the training process incorporated with incremental pruning, the model contains a set of weights with values set to zero.
From the point of view of this paper, the most important element of pruning is the extracting information the lowest sparsity level occurs with the \textit{K} output channels. This information is used to unify the sparsity level for each output channel. This procedure is significant for GPU implementation, where the execution time is depends upon the output channel with the lowest sparsity (see Section \ref{implemenation_desc}). Having a standardized sparsity level, pruned weights enables the compressed sparse row (CSR) format for each convolution layer to be built. To represent the matrix, the CSR format needs to build three arrays: 
\begin{itemize}
    \item \textit{Values} - these contains only non-zero elements. 
    \item \textit{Coldix} - on each position contains a offset for the value on equivalent position in the \textit{value} array. \textit{Park et al.} \cite{park2016faster}, proposed pre-computed value in the coldix matrix to store indexes from the input array which will be used to perform convolution. Thanks to this, there is no necessity to calculate these indexes during convolution which decrease calculation time.
    \item \textit{Rowptr} - \textit{rowptr[i]} is the point to the first non-zero element of the \textit{i}th output channel. Note that the result of \textit{rowptr[i+1] - rowptr[i]} is the number of non-zero elements in the \textit{i}th output channel. In our approach, this number is the same for each output channel thanks to the aforementioned standardized sparse level and we call this the $sparse\_level$. The only modification which must be done to avoid calculating the sparsity level separately for each output channel is mark some zero value as "non-zero" and during building, the CSR format treats them as normal value. This operation does not change the result and needs extra-memory. However, having the same sparsity for each output channel determines that each warp on the GPGPU has the same number of iterations, which is known before running the CUDA kernel which leads to the kernel's faster execution. These special "non-zero" values are chosen to not excessively jump thorough memory this mean there are choose zeros with side-by-side indices to guarantee contiguous direct memory access . 
\end{itemize}

\section{Convolution operation using a sparse operation on GPGPU}
\label{implemenation_desc}
\begin{figure}[ht]
  \centering\includegraphics[width=\linewidth, viewport=0 400 566 775,
  clip=true]{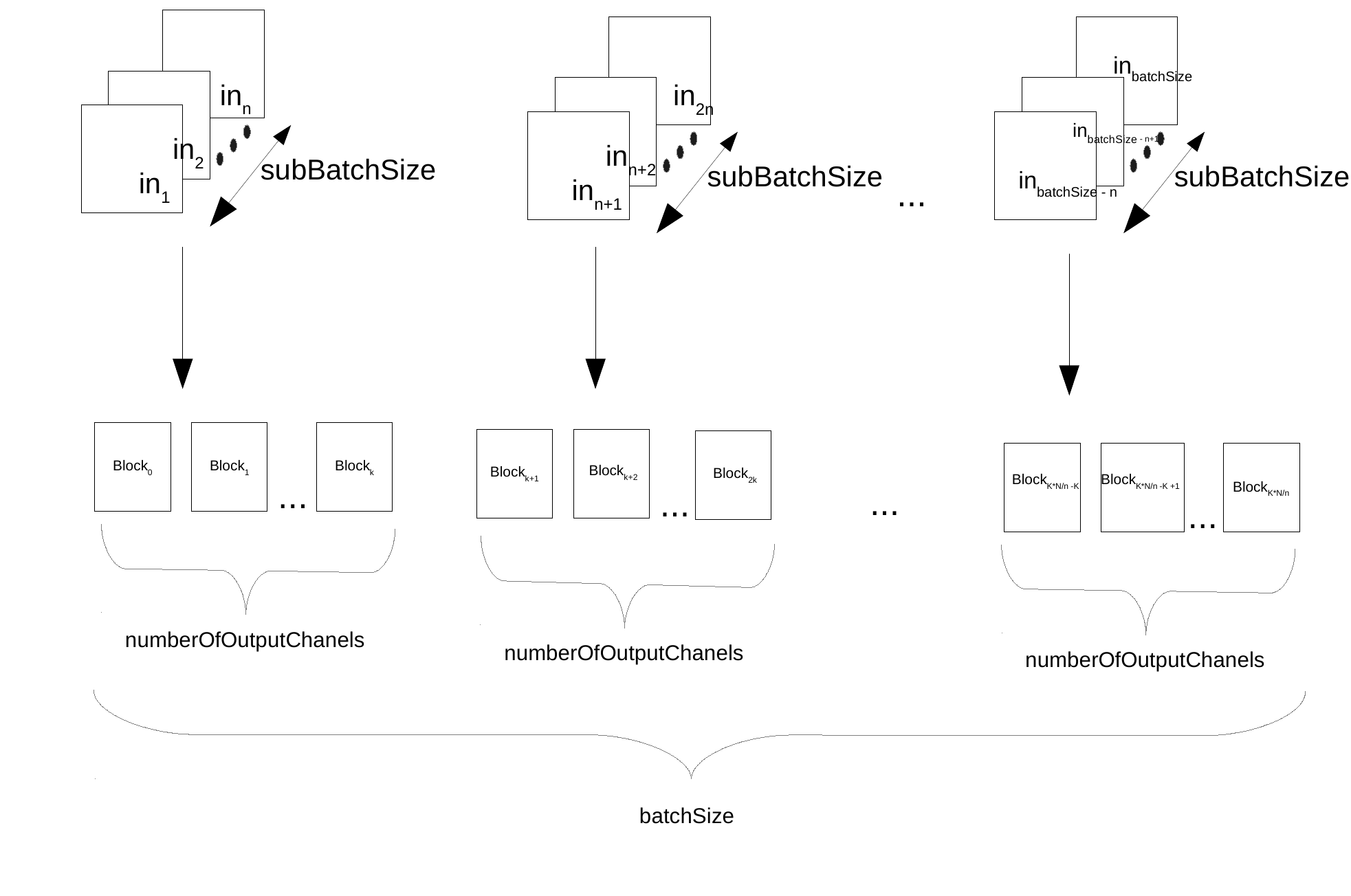}
    \caption{The total number of thread's block using to perform convolution for a single layer}
    \label{numberOfBlock_schema}
\end{figure}

\begin{figure}[ht]
  \centering\includegraphics[width=\linewidth, viewport=0 490 496 820,
  clip=true]{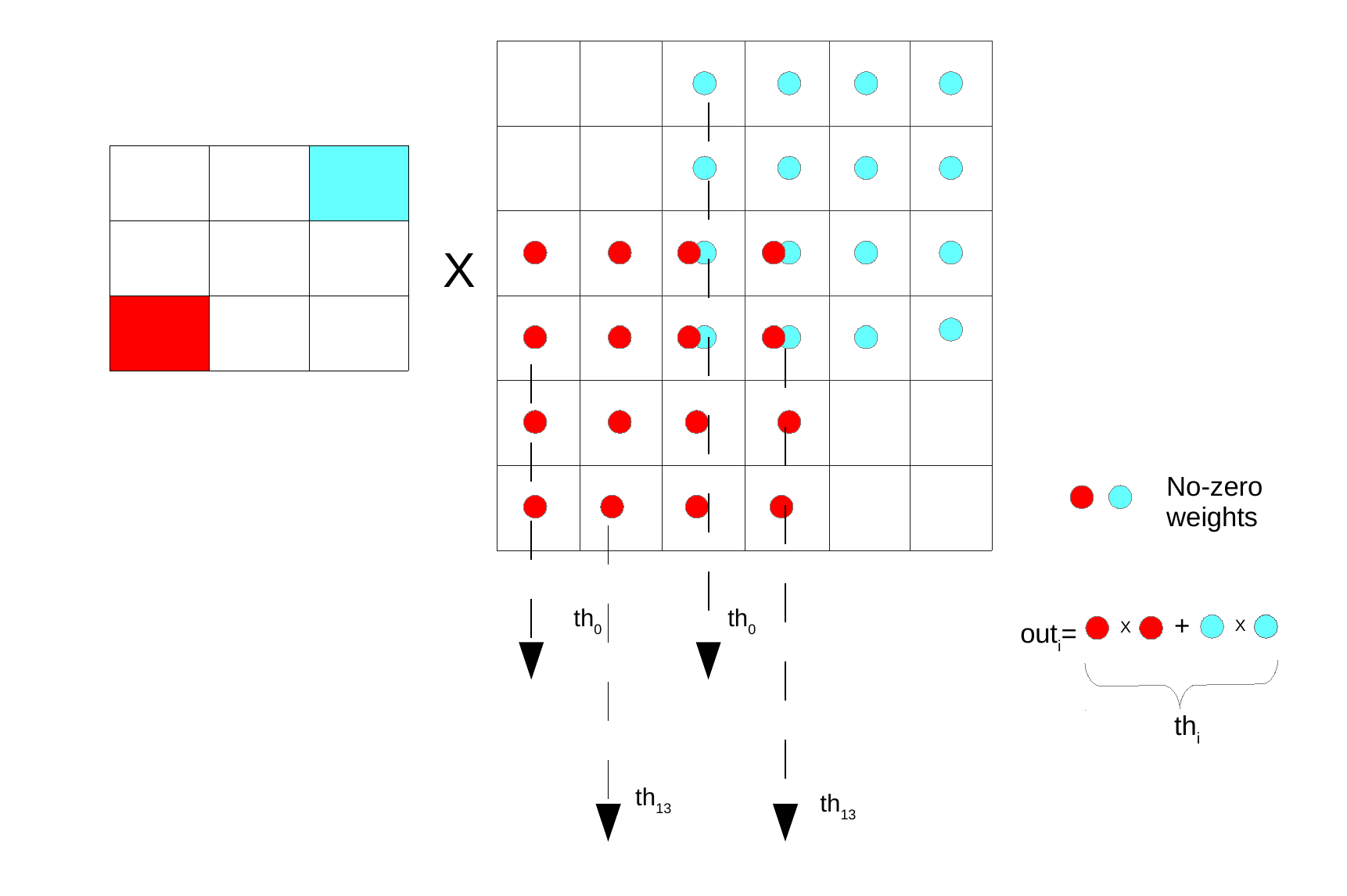}
    \caption{Calculating convolution using a sparse operation}
    \label{sparsce_conv_schema}
\end{figure}

To perform convolution by usage of a sparse operation \textit{the direct sparse convolution} \cite{park2016faster} was used. In parallel implementation we use an approach proposed by \cite{chen2018escoin}. The input data are stored in NCHW format (batch size, channel, height, width). The weights are stored in the CSR format where \textit{coldix} and \textit{value} arrays are loaded into shared memory. Each single thread block, calculates one output channel so for one input vector the total number of thread blocks is equal to the number of output channels. In our approach, we optimize the number of input vectors from the input batch, which will be handled by this number of blocks and is denoted as $subBatchSize$. As a result, the total number of thread blocks is equal to $\frac{batchSize*numberOfOutputChannel}{subBatchSize}$ which is presented by Fig. \ref{numberOfBlock_schema}. 
As it turns out, this number depends on the layer size and belongs to \{2, 4, 8\}  (see section \ref{result}). This optimal number is not the same for each type of layer due to the cache limitation and when this particular memory is missed, data are put into global memory which is very slow. For this reason, in this paper this number was experimentally fixed for each layer. During the calculation of the convolution, non-zero values from the \textit{values} array and pre-calculated indices of the input vector from the \textit{coldix} array are loaded from shared memory into the thread local memory and it is reused for $subBatchSize$ input vectors. This procedure enables maximum limitation of the reading from shared memory. Similar to the weights and indexes values, the partial sums are stored in registers and are copied to global memory after calculations. The number of threads used for the calculation of one output channel for one vector is determined by the output size of convolution. Each single thread is responsible for calculating one single output value by multiplication with the weight with corresponding input value, accumulating the partial sum and writing the final result to the global memory as is shown in Fig. \ref{sparsce_conv_schema}. The total number of working threads in a single threads block is determined by the sparsity level which in our approach is the same for each output channel and is equal to sparsity of channel with minimum value (see Section \ref{weigh_pruning}). 
In the version of the implementation where each channel has different sparsity the execution time was longest $\sim28\%$ for VGG-16 $3x3$ and $1x1$ convolution type and $\sim26\%$ in the case of both convolution layers from CNN-non static. As an improvements both weights and input feature maps are marked as constant in order to hold them in the L2 cache, and coalesced memory access is provided. This convolution function is chosen for the cuDnn flow and performs convolution instead of the cuDnn function in the case of specific sparsity level (higher than $\sim$ 90\% for vgg-16 and 1x1 convolution, and more than $\sim$78\% for CNN-non static) and this is achieved only for some layers, as is shown in the next sections. The greatest acceleration of direct sparse convolution over the cuDnn was achieved for the 1D convolution. In this case, the input data are in the shape of a vector;  therefore to preform convolution by usage of the \textit{direct sparse method}, less memory jumps are needed than with 2D convolution. 
Besides sparsity we are checking how precision reduction can accelerate the calculation of convolution in sparse implementation and with usage of dedicated libraries. In order to achieve this, data are transformed  from \textit{float} to \textit{half} type for both weights (the \textit{value} array) and input data. Cuda-Math-Api\footnote{https://docs.nvidia.com/cuda/cuda-math-api/} is used in order to perform calculations with half the precision on the GPU. Cuda-Math-Api provides transformations and mathematical functions for \textit{half} type. As described in \cite{10.1145/3007787.3001163} and \cite{motaz2020} the 16-bit half precision is sufficient to to keep the CNN models accuracy on the same level.

\begin{table}[]
\centering
\begin{tabular}{|c|c|c|l|l|}
\hline
\textbf{} & \textbf{\begin{tabular}[c]{@{}c@{}}Accuracy\end{tabular}} & \multicolumn{3}{c|}{{\textbf{\begin{tabular}[c]{@{}c@{}}Weighted sparsity\end{tabular}}}} \\
                                    &                                                                                & \multicolumn{3}{c|}{}                                                                                 \\ \hline
\textbf{VGG16 dense}                      & 90.01                                                                          & \multicolumn{3}{c|}{0.0}                                                                             \\ \hline
\textbf{VGG16 sparse}                       & 92.5                                                                          & \multicolumn{3}{c|}{96.2}                                                                             \\ \hline
\end{tabular}

\caption{\label{tab:vgg_pruned}Table with VGG16 pruning results on CIFAR10.}
\end{table}

\begin{table}[]
\centering
\begin{tabular}{|c|c|c|l|l|}
\hline
\textbf{} & \textbf{\begin{tabular}[c]{@{}c@{}}Accuracy\end{tabular}} & \multicolumn{3}{c|}{{\textbf{\begin{tabular}[c]{@{}c@{}}Weighted sparsity\end{tabular}}}} \\
                                    &                                                                                & \multicolumn{3}{c|}{}                                                                                 \\ \hline
\textbf{VGG16 dense}                      & 64.99                                                                          & \multicolumn{3}{c|}{0.0}                                                                             \\ \hline
\textbf{VGG16 sparse}                       & 68.31                                                                          & \multicolumn{3}{c|}{92.2}                                                                             \\ \hline
\end{tabular}

\caption{\label{tab:vgg_pruned_100} Table with VGG16 pruning results on CIFAR100.}
\end{table}

\begin{table}[]
\centering
\begin{tabular}{|c|c|}
\hline
{\textbf{}} & {\textbf{Accuracy}}  \\
                                                                     \hline
\textbf{VGG16 sparse 4b/16b}                      & 68.2                                                                                                                                                      \\ \hline
\textbf{VGG16 sparse 16b/16b}                       & 68.3                                                                                                                                                      \\ \hline
\end{tabular}

\caption{\label{tab:vgg_quant} Table with VGG16 quantization results on CIFAR100.}
\end{table}

\begin{table}[]
\centering
\begin{tabular}{|c|c|c|l|l|}
\hline
\textbf{} & \textbf{\begin{tabular}[c]{@{}c@{}}Accuracy\end{tabular}} & \multicolumn{3}{c|}{{\textbf{\begin{tabular}[c]{@{}c@{}}Weighted sparsity\end{tabular}}}} \\
                                    &                                                                                & \multicolumn{3}{c|}{}                                                                                 \\ \hline
\textbf{Resnet50 dense}                      & 92.3                                                                          & \multicolumn{3}{c|}{0.0}                                                                             \\ \hline
\textbf{Resnet50 sparse}                       & 94.1                                                                          & \multicolumn{3}{c|}{97.14}                                                                             \\ \hline
\end{tabular}

\caption{\label{tab:resnet_pruned} Table with Resnet50 pruning results on CIFAR10.}
\end{table}

\begin{table}[]
\centering
\begin{tabular}{|c|c|c|l|l|}
\hline
\textbf{} & \textbf{\begin{tabular}[c]{@{}c@{}}Accuracy\end{tabular}} & \multicolumn{3}{c|}{{\textbf{\begin{tabular}[c]{@{}c@{}}Weighted sparsity\end{tabular}}}} \\
                                    &                                                                                & \multicolumn{3}{c|}{}                                                                                 \\ \hline
\textbf{Resnet50 dense}                      & 67.06                                                                          & \multicolumn{3}{c|}{0.0}                                                                             \\ \hline
\textbf{Resnet50 sparse}                       & 78.23                                                                          & \multicolumn{3}{c|}{90.14}                                                                             \\ \hline
\end{tabular}

\caption{\label{tab:resnet_pruned_100}Table with Resnet50 pruning results on CIFAR100.}
\end{table}

\begin{table}[]
\centering
\begin{tabular}{|c|c|l|l|}
\hline
\textbf{} & \textbf{\begin{tabular}[c]{@{}c@{}}Accuracy\end{tabular}}                                                                              \\ \hline
\textbf{Resnet50 sparse 4b/16b}                      & 93.8                                                                                                                                                   \\ \hline
\textbf{Resnet50 sparse 16b/16b}                       & 94.0                                                                                                                                                     \\ \hline
\end{tabular}
\label{tab:resnet_quant}
\caption{Table with Resnet50 quantization results on CIFAR100.}
\end{table}

\begin{table}[]
\centering
\begin{tabular}{|c|c|c|l|l|}
\hline
\textbf{} & \textbf{\begin{tabular}[c]{@{}c@{}}Accuracy\end{tabular}} & \multicolumn{3}{c|}{{\textbf{\begin{tabular}[c]{@{}c@{}}Weighted sparsity\end{tabular}}}} \\
                                    &                                                                                & \multicolumn{3}{c|}{}                                                                                 \\ \hline
\textbf{DenseNet dense}                      & 82.0                                                                          & \multicolumn{3}{c|}{0.0}                                                                             \\ \hline
\textbf{DenseNet sparse}                       & 84.0                                                                          & \multicolumn{3}{c|}{89.5}                                                                             \\ \hline
\end{tabular}
\label{tab:densenet_pruned}
\caption{Table with DenseNet pruning results on CIFAR100.}
\end{table}

\begin{table}[]
\centering
\begin{tabular}{|c|c|c|l|l|}
\hline
\textbf{} & \textbf{\begin{tabular}[c]{@{}c@{}}Accuracy\end{tabular}}                                                                            \\ \hline
\textbf{DenseNet sparse 4b/16b}                      & 83.5                                                                                                                                                    \\ \hline
\textbf{DenseNet sparse 16b/16b}                       & 84.1                                                                                                                                                      \\ \hline
\end{tabular}
\label{tab:densenet_quant}
\caption{Table with DenseNet quantization results on CIFAR100.}
\end{table}

\begin{algorithm}
\begin{algorithmic}[1]
\REQUIRE{$sparsity\_thresholds\_for\_specific\_layers$}
\STATE{$run\_pruning$}
\STATE{$quantization$}
\STATE{$layers\_efficiency\_comparing(cudnn, direct)$}
\STATE{$network\_configuration$}
\caption{Main scheme applying reduced bit format and unstructured sparsity in GPU}
\label{alg:main}
\end{algorithmic}
\end{algorithm}

The described process of pruning produces layers with certain number of zero weights. Then these layers are mapped to direct sparse implementation. At the end quantization is applied. Finally efficiency of layers with reduced number of weights and bit format is compared with CuDnn. Based on these results network is configured to be partially run with direct sparse 
approach. The whole process is described in alg. \ref{alg:main}.

\section{Results}
\label{result}

\begin{figure}[ht]
  \centering\includegraphics[scale=0.3]{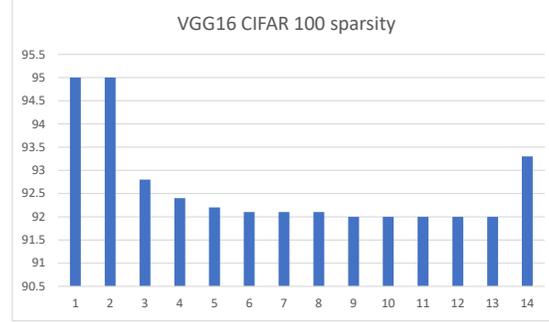}
    \caption{VGG16 on CIFAR100}
    \label{fig:vgg16_cifar100}
\end{figure}

\begin{figure}[ht]
  \centering\includegraphics[scale=0.3]{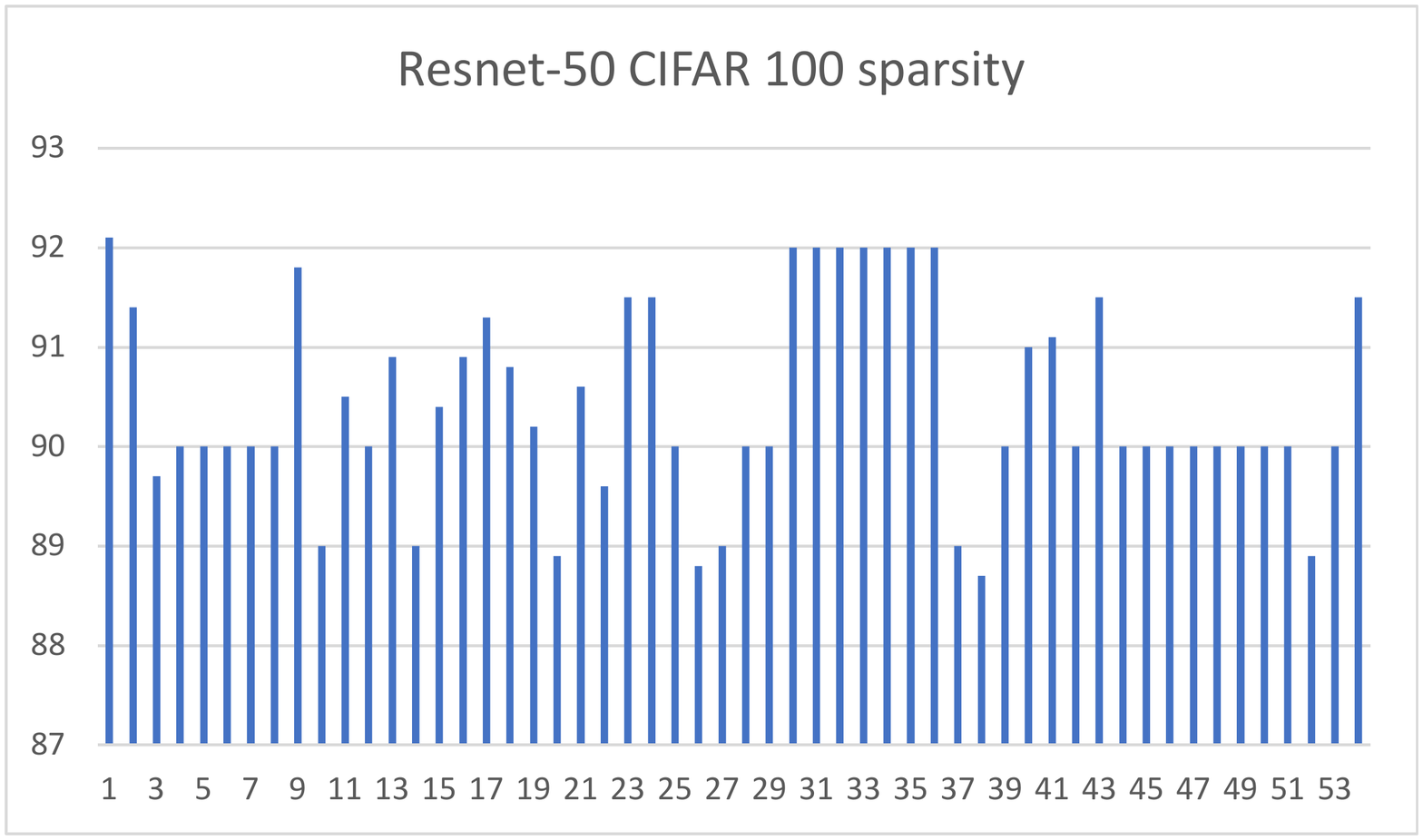}
    \caption{Resnet50 on CIFAR100}
    \label{fig:resnet50_cifar100}
\end{figure}

In tables \ref{tab:vgg_pruned}, \ref{tab:vgg_pruned_100},  \ref{tab:resnet_pruned} and \ref{tab:resnet_pruned_100} the accuracy and level of sparsity are described which were achieved by pruning algorithm on VGG16 and Resnet50 on CIFAR10 and CIFAR100. The dense models are the baseline models. We can see that apart from high sparsity and huge memory reduction the accuracy is increased. In case of Resnet50 it is worth to note that achieved pruned version is one of the smallest model in TOP40 models in the CIFAR100 ranking \cite{cifar100}. The results were achieved by running 200 epochs of the training process. In results only weighted sparsity are given. In case of Resnet50 about half of all layers are above speedup threshold (\textgreater 90\%) and in VGG16 all except the first layer (fig. \ref{fig:vgg16_cifar100} and \ref{fig:resnet50_cifar100}). In tables \ref{tab:vgg_quant} and \ref{tab:resnet_quant} the results of quantization approaches are presented. The linear 16 bit (half precision both for weights and activations - 16b/16b) quantization and cluster based 4 bit quantization were performed (weights are mapped to 16 centroids, each centroid represented in 16 bit format). The quanization was run on a pretrained and pruned models. It is worth noting that in all cases no drop in accuracy was observed (<0.5\%). 
In tables \ref{tab:densenet_pruned} and \ref{tab:densenet_quant} results for DenseNet model are described for the same pruning and quantization configurations. Improvement in accuracy after pruning was observed and slight drop after quantization. 

\begin{table*}[]
\centering
\resizebox{\textwidth}{!}{\begin{tabular}{|c|c|c|c|c|}
\hline
\textbf{\begin{tabular}[c]{@{}c@{}}Convolution size\\  (CHWK)\end{tabular}} & \textbf{Escoin time - float} & \textbf{\begin{tabular}[c]{@{}c@{}}cuDnn time - float\\ \textbackslash{}conv algorithm\end{tabular}} & \textbf{Escoin time - half} & \textbf{\begin{tabular}[c]{@{}c@{}}cuDnn time -half\\ \textbackslash{}conv algorithm\end{tabular}} \\ \hline
3x224x224x64                                                                & 2.48                                 & 2.82\textbackslash{}GEMM                                                                                & 2.21                                & 2.71\textbackslash{}GEMM                                                                               \\ \hline
64x224x224x64                                                               & 60.73                                & 19.07\textbackslash{}WINOGRAD                                                                               & 27.08                               & 31.82\textbackslash{}GEMM                                                                              \\ \hline
64x112x112x128                                                              & 16.45                                & 10.56\textbackslash{}WINOGRAD                                                                           & 8.87                                & 9.48\textbackslash{}GEMM                                                                               \\ \hline
128x112x112x128                                                             & 28.11                                & 17.28\textbackslash{}WINOGRAD                                                                           & 17.31                               & 15.88\textbackslash{}GEMM                                                                              \\ \hline
128x56x56x256                                                               & 13.61                                & 9.21\textbackslash{}FFT-TILING                                                                          & 8.74                                & 7.81\textbackslash{}GEMM                                                                               \\ \hline
256x56x56x256                                                               & 23.72                                & 14.27\textbackslash{}FFT-TILING                                                                         & 15.83                               & 16.09\textbackslash{}GEMM                                                                              \\ \hline
256x28x28x512                                                               & 9.34                                 & 6.72\textbackslash{}FFT-TILING                                                                          & 6.07                                & 7.86\textbackslash{}GEMM                                                                               \\ \hline
512x28x28x512                                                               & 16.06                                & 15.02\textbackslash{}FFT-TILING                                                                         & 14.01                               & 16.82\textbackslash{}GEMM                                                                              \\ \hline
512x14x14x512                                                               & 4.31                                 & 4.84\textbackslash{}FFT-TILING                                                                          & 4.18                                & 4.66\textbackslash{}GEMM                                                                               \\ \hline
\end{tabular}}
\caption{\label{tab:vgg_16_table}Time results [ms] for VGG-16 (sparsity $\sim90\%$)}
\end{table*}

\begin{table*}[]
\centering
\begin{tabular}{|c|c|c|c|c|c|c|}
\hline
\textbf{Layer name}                                                                                   & \textbf{\begin{tabular}[c]{@{}c@{}}Sparsity\\ {[}\%{]}\end{tabular}} & \textbf{\begin{tabular}[c]{@{}c@{}}subBatchSize\\ (optimal)\end{tabular}} & \textbf{\begin{tabular}[c]{@{}c@{}}Escoin \\ -float\end{tabular}} & \textbf{\begin{tabular}[c]{@{}c@{}}Cudnn\\ -float\end{tabular}} & \textbf{\begin{tabular}[c]{@{}c@{}}Escoin\\ -half\end{tabular}} & \textbf{\begin{tabular}[c]{@{}c@{}}Cudnn\\ -half\end{tabular}} \\ \hline
\textbf{\begin{tabular}[c]{@{}c@{}}densenet121/\\ dense\_block\_3/\\ dense\_layer24\end{tabular}}     & 87,5                                                                 & 8                                                                         & 0.69                                                              & 0.73                                                            & 0.62                                                            & 0.68                                                           \\ \hline
\textbf{\begin{tabular}[c]{@{}c@{}}densenet\_121/\\ dense\_block\_3/\\ dense\_layer\_24\end{tabular}} & 91                                                                   & 8                                                                         & 0.46                                                              & 0.73                                                            & 0.41                                                            & 0.68                                                           \\ \hline
\textbf{\begin{tabular}[c]{@{}c@{}}densenet161/\\ dense\_block\_4/\\ dense\_layer\_16\end{tabular}}   & 91                                                                   & 4                                                                         & 2.43                                                              & 2.56                                                            & 2.37                                                            & 2.49                                                           \\ \hline
\textbf{\begin{tabular}[c]{@{}c@{}}densenet161/\\ dense\_block\_3/\\ dense\_leayer\_16\end{tabular}}  & 93                                                                   & 4                                                                         & 1.81                                                              & 2.56                                                            & 1.75                                                            & 2.49                                                           \\ \hline
\end{tabular}
\caption{Time result [ms] for given layer from DenseNet model for 1x1 convolution}
\label{tab:dense}
\end{table*}

\begin{table}[]
\centering
\begin{tabular}{|c|c|c|l|l|}
\hline
\textbf{Data type} & \textbf{\begin{tabular}[c]{@{}c@{}}CUDNN Time  \end{tabular}} & \multicolumn{3}{c|}{{\textbf{\begin{tabular}[c]{@{}c@{}}Escoin \\ time\end{tabular}}}} \\
                                    &                                                                                & \multicolumn{3}{c|}{}                                                                                 \\ \hline
\textbf{float}                      & 0.35                                                                          & \multicolumn{3}{c|}{0.32}                                                                             \\ \hline
\textbf{half}                       & 0.30                                                                         & \multicolumn{3}{c|}{0.27}                                                                             \\ \hline
\end{tabular}
\caption{Time results [ms] for ResNet. 256 filters 1x1x64  (sparsity $\sim90\%$)}
\label{tab:res_net_64}
\end{table}

\begin{table}[]
\centering
\begin{tabular}{|c|c|c|l|l|}
\hline
\textbf{Data type} & \textbf{\begin{tabular}[c]{@{}c@{}}CUDNN Time\end{tabular}} & \multicolumn{3}{c|}{{\textbf{\begin{tabular}[c]{@{}c@{}}Escoin time\end{tabular}}}} \\
                                    &                                                                                & \multicolumn{3}{c|}{}                                                                                 \\ \hline
\textbf{float}                      & 0.37                                                                          & \multicolumn{3}{c|}{0.31}                                                                             \\ \hline
\textbf{half}                       & 0.29                                                                        & \multicolumn{3}{c|}{0.24}                                                                             \\ \hline
\end{tabular}
\caption{Time results  [ms] for ResNet. 64 filters 1x1x256  (sparsity $\sim90\%$)}
\label{tab:res_net_256}
\end{table}

\begin{table}[]
\centering
\begin{tabular}{|c|c|c|c|c|}
\hline
\textbf{Data type} & \textbf{\begin{tabular}[c]{@{}c@{}}CUDNN Time \end{tabular}} & \multicolumn{3}{c|}{\textbf{Escoin time for given sparsity}} \\ \cline{3-5} 
                                    &                                                                                & \textbf{77\%}      & \textbf{83\%}     & \textbf{87,5\%}     \\ \hline
\textbf{float}                      & 0.192                                                                         & 0.176              & 0.126             & 0.102               \\ \hline
\textbf{half}                       & 0.161                                                                         & 0.145              & 0.097             & 0.069               \\ \hline
\end{tabular}
\caption{Time results  [ms] for CNN-non-static for input 300x64. kernel size 2}
\label{tab:cnn_non_static_2}
\end{table}

\begin{table}[]
\centering
\begin{tabular}{|l|l|l|c|c|}
\hline
\textbf{Data type} &  \textbf{\begin{tabular}[c]{@{}l@{}}CUDNN Time\\ \end{tabular}} & \multicolumn{3}{l|}{\textbf{Escoin time for given sparsity}} \\ \cline{3-5} 
                                    &                                                                                & \textbf{77\%}      & \textbf{83\%}     & \textbf{87,5\%}     \\ \hline
\textbf{float}                      & \multicolumn{1}{c|}{0.231}                                                     & 0.236              & 0.188             & 0.135               \\ \hline
\textbf{half}                       & \multicolumn{1}{c|}{0.204}                                                      & 0.182              & 0.148             & 0.103               \\ \hline
\end{tabular}
\caption{CNN-non-static for input 300x64. kernel size 3}
\label{tab:cnn_non_static_3}
\end{table}

All of the presented calculations were performed on the Nvidia Tesla V100-SXM2-32GB\footnote{https://www.nvidia.com/en-us/data-center/v100/}. The batch size is always equal 128 (for others or 64 or 256, the proportions are the same) and the final execution time is calculated as the average of 10 iterations. In our experiments, we let the cuDnn library choose the algorithm which would be used to perform convolution for different layers and data types. The tables below, presents the algorithm which was used by the cuDnn library in addition to the execution time for each layer. For each experimental calculation of convolution the $subBatchSize$ was determined (see Section \ref{implemenation_desc}) which for last three layers from VGG-16, for $1\times1$ convolution layers from ResNet50 is 8. For the remaining VGG-16 and some layers and two and for CNN-non static with filter size two, this value is 4, and for CNN-non static with filter size three, this parameter is 2. For DenseNet model this value depend on layer is 4 or 8 (see table \ref{tab:dense}). The presented results were measured with the optimal value of this parameter. Without setting this value by the method proposed in this paper, it would not be possible to achieve better performance than cuDnn because when the number of block is equal to $numberOfOutputChannel*batchSize$, for VGG-16 and $1x1$ layers, the performance decreased by around $\sim10\%$. In the case of CNN-non static, the decrease was $\sim12\%$. An even larger drop in performance occurred when all the data from the batch was processed by $K$ blocks this value was between $\sim38\%$ and $\sim45\%$. Table \ref{tab:vgg_16_table} includes the results of time execution for the VGG-16 model model, where for each layers the sparsity was set at $\sim90\%$ because this is the lowest sparsity level for which the \textit{direct sparse convolution} algorithm is more effective than the cuDnn library. Despite such a high sparse level, the improvement over the cuDnn was not achieved for every layer. Only for the first and last three layers where the input size in NCHW format, is $3\times226\times226$ and $512\times16\times16$ respectively, was the improvement gained for \textit{float} ($\sim13\%$-first layer and $\sim12\%$-last layer) and \textit{half} ($\sim22\%$-first layer and $\sim11\%$) data type. In addition, in both cases the algorithm is faster for the \textit{half} type which is not obvious for the cuDnn library, where for \textit{half-precision}, the cuDnn always performs convolution by the \textit{GEMM} algorithm. This way of calculating the convolution on \textit{half} type, for the VGG-16's convolution layers with input sizes $64\times226\times226$, $256\times58\times58$ and $512\times30\times30$ is less effective than performing this on \textit{float} type with the use of \textit{FFT} or \textit{WINOGRAD} algorithm. Taking into account only the data in \textit{half} type format, the sparse approach can additionally improve performance of the VGG-16's conv layers with the follow input size: $64\times114\times114$, $128\times58\times58$, $256\times30\times30$, $512\times30\times30$. Having the same level of sparsity as in the VGG-16 architecture, there is the possibility to achieve better performance than the cuDnn for the $1\times1$ convolution in Resnet architecture. For this type of convolution, the cuDnn always uses the \textit{GEMM} algorithm and the result for this are included in Tables \ref{tab:res_net_64} and \ref{tab:res_net_256}. Table \ref{tab:dense} contains time result for the same type of convoltuion from DenseNet architecture. In this case the sparsity level and \textit{subBatchSize} were determined to achieve the best performance for particular layer. The most effective performance of the \textit{direct sparse convolution} method is achieved for the 1D convolution which is dedicated to the time series data. A significant acceleration compared to cuDnn was reached for the CNN-non static, where for convolution layer with kernel size 2, the sufficient sparsity level is $\sim77\%$ to gain a $\sim9\%$ and $\sim11\%$ speed increase for the \textit{float} and \textit{half} data types, respectively (see Tables \ref{tab:cnn_non_static_2} and \ref{tab:cnn_non_static_3}). In tab.\ref{tab:vgg_real} real times of pruned VGG-16 (\ref{tab:vgg_pruned}) are described. The speedup from few to several percent to cuDnn can be observed (see tab.\ref{tab:vgg_16_table}).



\begin{table}[]
\centering
\begin{tabular}{|c|c|c|c|}
\hline
\textbf{\begin{tabular}[c]{@{}c@{}}Convolution size\\ (CHWK)\end{tabular}} & \textbf{\begin{tabular}[c]{@{}c@{}}Sparsity\\ {[}\%{]}\end{tabular}} & \textbf{\begin{tabular}[c]{@{}c@{}}Escoin time\\  - float\end{tabular}} & \textbf{\begin{tabular}[c]{@{}c@{}}Escoin time\\  - half\end{tabular}} \\ \hline
64x224x224x64                                                              & 90                                                                   & 60.73                                                                           & 27.08                                                                          \\ \hline
64x112x112x128                                                             & 92                                                                   & 15.97                                                                           & 8.64                                                                           \\ \hline
128x112x112x128                                                            & 93                                                                   & 27,12                                                                           & 16.22                                                                          \\ \hline
128x56x56x256                                                              & 92                                                                   & 12.28                                                                           & 8.42                                                                           \\ \hline
256x56x56x256                                                              & 90.8                                                                 & 21.81                                                                           & 14.23                                                                          \\ \hline
256x28x28x512                                                              & 92                                                                   & 9.11                                                                            & 5.92                                                                           \\ \hline
512x28x28x512                                                              & 92                                                                   & 15.28                                                                           & 13.67                                                                          \\ \hline
512x14x14x512                                                              & 92                                                                   & 4.23                                                                            & 4.08                                                                           \\ \hline    
\end{tabular}
\caption{Escoin time  [ms] for VGG-16 with real sparsity}
\label{tab:vgg_real}
\end{table}

\section{Conclusions and future work}
This work is focused on speeding up the convolution operation on GPGPU through the use of the sparse matrix operation and the representation of data at a reduced level of precision. In particular, this strategy makes maximum use of knowledge about the number of produced zero values as a result of the pruning process. The time results obtained from the proposed solution are comparable with the convolution kernel from the cuDnn library, which is recognized as the most effective way to perform convolution on GPGPUs. We have presented concrete cases when it is worth performing convolution using the \textit{direct sparse convolution} in the cuDnn place. The most improvements are archived for 1D convolution because for this type, the cuDnn library always chooses the GEMM method to perform convolution which does not provide such a strong performance such as the WINOGRAD or the FFT-TILIING method which are used to performing 2D convolution. It is shown that 2D convolution using \textit{direct sparse convolution} can also outperform cuDnn algorithms. Additionally, we have examined the influence conducting the calculation using reduced precision on time efficiency. The next contribution is showing speedup of DL models on real examples using pruning. It is worth to mention that pruning can significantly improve the accuracy of DL big models when they are used on less complex and reduced datasets.
The future works will concentrate on different models on CIFAR100, pruning models based on transfer learning and models for object detection and image segmentation. The pruning will be explored more to check if it is possible to further increase the sparsity. The next approach will adaptation of pruned models in FPGA.

\nocite{*} 
\bibliographystyle{unsrt}
\bibliography{bibliograph}
\end{document}